\documentclass[journal,twoside,web]{ieeecolor}
\usepackage{generic}
\usepackage{cite}
\usepackage[nosumlimits,nonamelimits]{amsmath}
\usepackage{amssymb,amsfonts}
\usepackage{algorithmic}
\usepackage{graphicx}
\usepackage{multirow}
\usepackage{booktabs}
\usepackage{textcomp}
\usepackage{orcidlink}
\usepackage{subcaption}
\pdfoutput=1
\def\BibTeX{{\rm B\kern-.05em{\sc i\kern-.025em b}\kern-.08em
    T\kern-.1667em\lower.7ex\hbox{E}\kern-.125emX}}
\begin{document}
\title{Subtyping patients with chronic disease using longitudinal BMI patterns}
\author{Md Mozaharul Mottalib \orcidlink{0000-0003-4930-0365}, Jessica C Jones-Smith \orcidlink{0000-0001-8962-1695}, Bethany Sheridan, and Rahmatollah Beheshti \orcidlink{0000-0001-8912-3063}, 

\thanks{Md Mozaharul Mottalib (mmmdip@udel.edu) is with the University of Delaware, Newark, DE, USA. Jessica C Jones-Smith is with the University of Washington, Seattle, WA, USA. Bethany Sheridan was with athenahealth Inc., Watertown, MA, USA. at time of working on this. Rahmatollah Behesht (rbi@udel.edu)i is with the University of Delaware.}
\thanks{Our study was approved by the University of Delaware institutional review board (1627409). The study was supported by the NIH awards 3P20GM103446 and 5P20GM113125, as well as Robert Wood Johnson Foundation’s award 76778.}}
\maketitle

\begin{abstract}
Obesity is a major health problem, increasing the risk of various major chronic diseases, such as diabetes, cancer, and stroke. While the role of obesity identified by cross-sectional BMI recordings has been heavily studied, the role of BMI trajectories is much less explored. In this study, we use a machine learning approach to subtype individuals’ risk of developing 18 major chronic diseases by using their BMI trajectories extracted from a large and geographically diverse EHR dataset capturing the health status of around two million individuals for a period of six years. We define nine new interpretable and evidence-based variables based on the BMI trajectories to cluster the patients into subgroups using the k-means clustering method. We thoroughly review each cluster’s characteristics in terms of demographic, socioeconomic, and physiological measurement variables to specify the distinct properties of the patients in the clusters. In our experiments, the direct relationship of obesity with diabetes, hypertension, Alzheimer’s, and dementia has been re-established and distinct clusters with specific characteristics for several of the chronic diseases have been found to be conforming or complementary to the existing body of knowledge.
\end{abstract}

\begin{IEEEkeywords}
patient subtyping,  interpretable machine learning, BMI trajectories, obesity, chronic diseases
\end{IEEEkeywords}

\section{Introduction}
\label{sec:introduction}
\IEEEPARstart{O}{besity} is known as an important risk factor for major chronic diseases, such as type 2 diabetes \cite{RN117, RN139}, different cancer types \cite{RN97, RN143} and cardiovascular diseases \cite{RN145}. Over the past few decades, the prevalence of obesity has rapidly increased in almost every part of the world, with an estimated 13.24\% of the world \cite{RN146} and 39.8\% of the US population having obesity in 2016 \cite{RN147}. While the relationships between obesity and other major diseases have been extensively studied in existing literature \cite{RN117, RN139, RN145, RN147, RN131, RN133, RN114, RN115, RN143, RN124, RN148, RN126, RN129, RN142, RN122}, subtyping individuals based on their longitudinal obesity patterns have not been studied, especially on a large scale. There is a growing body of evidence in the field of obesity research, as well as other epidemiology areas, showing the importance of considering historical obesity patterns in studying the effects of obesity \cite{RN116}. 

This study examines the hypothesis that longitudinal obesity patterns alone can predict individuals’ future patterns of chronic diseases. Specifically, our study aims at subtyping individuals’ risk of developing 18 major chronic diseases, by using only their obesity patterns as indicated by their BMI (body mass index) trajectories. While BMI is known to have some limitations in accurately measuring obesity \cite{RN235}, it is the standard, easily measurable, and widely adopted criterion that is being used to study obesity worldwide \cite{RN93, RN97, RN94}. In our work, we use a large EHR dataset capturing the health status of around two million individuals for a period of six years and define 18 separate cohorts, one for each studied chronic disease. We define nine new features based on individuals’  BMI trajectories capturing the existing findings in the field of obesity research. These interpretable features capture the temporal obesity differences across the individuals and are used to cluster each cohort. Using these new features, we then identify the subtypes (clusters) that show distinguishable patterns of disease incidence, and further study each identified subtype using the demographic, socioeconomic, and physiological measurement characteristics of its members. In addition to studying each of the 18 cohorts related to chronic diseases, we study a combined cohort (containing individuals with any of the 18 diseases), by  using a similar approach. Specifically, the primary contributions of our study are:

\begin{itemize}
\item {We define a series of new interpretable features based on longitudinal BMI values, informed by the current pieces of evidence in obesity research, to capture various temporal patterns of the recorded BMI values. This makes our approach especially unique, as traditional predictive methods (such as regression-based methods) generally do not capture temporal patterns, and more advanced methods (such as deep learning methods) often lack interpretability.}
\item {We evaluate the relevance of the features and show that the features capturing temporal BMI patterns are directly associated with obesity, diabetes, cancer, benign prostate hyperplasia, hip fracture, osteoporosis, hypertension, stroke, and dementia.}
\item {We define a series of patient subtypes using BMI trajectories and discuss the differences and disparities in each identified subtype with respect to demographic and socioeconomic factors, as well as physiological measurement values. }
\end{itemize} 

\section{Literature Review}
\textbf{Obesity and BMI trajectories.} Most studies that have looked at the relationship between body weight or BMI and other chronic diseases use single baseline (cross-sectional) body weights or BMI measurements \cite{RN114, RN110, RN113}. Some of these studies have shown a significant correlation between the history of obesity and the risk for later-life comorbidities\cite{RN117, RN145, RN133, RN114, RN124, RN148, RN129, RN142, RN122}. As an example of such studies, Stokes et. al has demonstrated that among the non-pregnant adults without obesity who had a previous history of obesity, there was an increased prevalence of eight obesity-related diseases \cite{RN110}. They showed that the odds ratio for each of the eight comorbid diseases was also shown to be significantly higher among individuals with a history of obesity than individuals not having any history of it \cite{RN110}. Some related studies focused on the time of developing obesity and the future risk of morbidity and showed that those who experienced early and rapid weight gain during early adulthood were most likely to develop hypertension, type 2 diabetes and arthritis in middle age \cite{RN112}. It has been also shown that weight gain from early through middle adulthood is associated with a significantly increased risk for type 2 diabetes, cancers, and cardiovascular diseases \cite{RN111}. 
Recent studies have also underlined the significance of using the historical body-weight patterns for forecasting the risks of future adverse health outcomes \cite{RN114, RN115, RN110, RN112, guo2021predicting}. For instance, the maximum body weight, defined as the highest BMI during a weight history period, has been at the center of attention for some time now. In a study by Yu et al. \cite{RN113},  those with the maximum BMI in the overweight, class 1 obesity (30$\leq$BMI$<$35), or class 2 obesity (35$\leq$BMI$<$40) categories had a risk proliferation for all-cause mortality as well as cause-specific mortality, compared to those maintaining a normal weight over 16 years. In another study, a direct association between the maximum BMI over 24 years of weight history and all-cause mortality across class 1 and 2 obesity categories compared with normal weight was found \cite{RN114}. Stokes et. al studied the association between excess weight and mortality using data from the National Health and Nutrition Examination Survey (NHANES) among adults aged 50 to 74 \cite{RN115} and showed an increase in disease prevalence and mortality with the increase in maximum weight or BMI.
Attard et al. \cite{RN167} assessed the risk of cardiovascular diseases by studying the BMI trajectories from adolescence to adulthood. Zheng et al. \cite{RN168} addressed the heterogeneity and mortality risk among the older population using BMI trajectories of adults aged 51 to 77 years from the US health and retirement study (1992-2008). Being overweight or obese at the onset of adulthood and BMI trajectories over the lifespan that result in obesity lead to an increased  risk of colorectal cancer  in populations who were overweight or obese at the onset of adulthood and had BMI trajectories that resulted in obesity \cite{RN169}. Obesity at age 50 years was associated with an increased risk of dementia in another study \cite{RN170}. These studies hold testaments for demonstrating the impact of body weight gain in predicting the risk of chronic diseases.

\textbf{Patient Subtyping.} Subtyping is a common method in epidemiological studies, closely related to stratification and phenotyping. For subtyping, unsupervised clustering methods are commonly used. To name a few representative studies consider the work by Masud et al, where they compute patient similarity to develop a clinical decision support system, where both the time series and static data were utilized to blend clustering \cite{RN109}. For only temporal clustering, an autoencoder-based end-to-end learning framework has been proposed in \cite{RN134}, where Deep Temporal Clustering (DTC) algorithm integrated dimensionality reduction and temporal clustering. For irregular time series, Bahadori et al. \cite{RN135} introduced a data augmentation technique using deep neural networks optimized for several clinical prediction tasks. Among the related work to study obesity, in one study, distinct subgroups of men and women with obesity were observed when a two-step cluster analysis was performed on a representative sample of American adults with obesity \cite{RN133}. Here, similar participants were clustered together by running an initial pre-clustering followed by a hierarchical clustering method. In another study on patients with obesity, three types of clusters were identified based on the number of comorbidities and the percentage of patients suffering from them \cite{RN131}. A sparse $k$-means algorithm was also used at two levels to identify subgroups among the groups of individuals based on 14 comorbidities related to obesity and contrasted them against experts’ annotations \cite{RN131}.  

\textbf{Feature Extraction.} Where feature selection is defined as a technique to reduce the number of features in the model by removing irrelevant or redundant data, feature extraction aims to combine or transform features into a different knowledge base. Previously we discussed that most epidemiological studies examining the association between BMI and comorbidities tend to focus on BMI measurement at one point in time only. Tracking trajectory patterns over time accounts for dynamic changes in the measurements. Multiple approaches have been used to explore trajectories. In a broader sense, interpreting time series in the form of features have been studied by researchers for years \cite{muller2021feature, mattsson2019group, niu2020developing}. Fulcher et al. \cite{fulcher2017hctsa} developed a technique called hctsa to extract features from time series to a set of over 7,700 features that each encode a different scientific analysis method. Lubba et al. \cite{lubba2019catch22} showed their method catch22 tailors the feature set filtering the outcome of the hctsa \cite{fulcher2017hctsa} into a set of 22 characteristics. Karasu et al. \cite{karasu2019recognition} used moving averages to generate trend lines and eventually extract features from them. In a study by Wen et al. \cite{wen2012childhood}, childhood BMI trajectory is modeled with statistical methods to form a set of interpretable feature that estimates the trajectory characteristics.

\section{Materials and Methods}

\renewcommand{\arraystretch}{1.15}
\begin{table}[t]
\caption{Our study variables. Disease prior status (patient-level) and incidence (visit-level)  are not shown below. }
\resizebox{\columnwidth}{!}{
\begin{tabular}{lll}
\toprule
\textbf{Variable} & \textbf{Range or Categories}
\\ \midrule
 Age & 20–70+ years (undefined for anonymity) \\ 
                               Gender                     & Male,   Female                                                                \\ 
                               Race/ethnicity             & White,   Hispanic/Latino, Black or African American, Other\textsuperscript{*}                                                                       \\ 
                               Insurance   type           & Commercial,   Medicaid, Medicare, Other, Self-pay\textsuperscript{†}     \\ 
                               Residence   area           & Metro,   Metro-adjacent, Rural\textsuperscript{‡} \\ 
                               Income   category          & High,   Low, Medium\textsuperscript{§}                           \\ \hline
 LDL   Cholesterol          & 44   – 189 mg/dL                                                                             \\ 
                               HbA1c                      & 5.1\%   – 11.7\%                                                                           \\ 
                               Systolic   blood pressure  & 98   – 166 mmHg                                                                                                               \\ 
                               Diastolic   blood pressure & 58   – 100 mmHg                                                                                                              \\ \bottomrule 
\end{tabular}
}
\tiny
\textsuperscript{*}Census categories. \textsuperscript{†}Based on the   preponderance of claims assigned to a patient during the study window. \textsuperscript{‡}Based on patient county   defined by the US Dept. of Agriculture (USDA). \textsuperscript{§}Based on median   household income in patient’s zip code from American Community Survey.
\label{table:1}
\end{table}

\subsection{Data Processing}
In this study, we have created a series of 18 cohorts, using a dataset of patients in the athenahealth system, a large provider of network-enabled services for hospital and ambulatory clients in the US. Patients visiting athenahealth providers are broadly representative of the nation’s outpatient visits when compared to national benchmarks provided by the National Ambulatory Medical Care Survey (NAMCS) \cite{RN153}. The study was approved by a local institutional review board at the University of Delaware (Ref\#: 1627409). The data was collected throughout a 6-year period from Jan 1, 2013, to Dec 31, 2018. Patients aged 20 or over with at least one visit to a primary care provider at a primary care practice were included in the cohort. The initial cohort contained 1,997,007 patients with around 14,473,471 visit records. We removed the patients with only one visit over the 6-year time to ensure all patients have a BMI trajectory. This reduced the number of patients to 1,531,374.
The dataset contained patient-level (static) and visit-level (dynamic) variables  (Table \ref{table:1}). Patient-level variables included age, gender, race/ethnicity, insurance type, residence, and income category. The visit-level variables contained the visit time (visit year and -month) and the physiological measurements (lab tests). We removed the samples with missing values at any of the variables. This way the patient-level data contained a time-labeled multivariate sequence of observations. 

We then defined 18 different cohorts (not  exclusive) comprised of individuals having a diagnosis of one of the 18 major chronic diseases, as well as an equal number of randomly chosen healthy controls (not having any of the 18 diseases). The 18 major chronic diseases included (Table \ref{table:2}): 1) hypothyroidism, 2) stroke, 3) Alzheimer’s or dementia, 4) anemia, 5) asthma, 6) heart failure or ischemic heart disease or acute myocardial infarction (AMI), 7) benign prostatic hyperplasia (BPH), 8) chronic kidney disease (CKD), 9) cancer (breast, colon, prostate, endometrial, or lung), 10) depression, 11) diabetes, 12) hip or pelvic fracture or osteoporosis, 13) hyperlipidemia, 14) hypertension, 15) obesity, 16) rheumatic arthritis, or osteoarthritis, 17) atrial fibrillation (AFib), 18) Chronic Obstructive Pulmonary Disease (COPD). These diseases are captured in our dataset as the prior conditions (patient-level, referring to the presence of the condition prior to the first recorded visit), as well as the incidence in each visit.  A patient having a disease when the diagnosis is present for that disease in more than 75\% of the visits is labeled as 1, and 0 otherwise. 

\begin{table}[ht]
\tiny
\caption{Cohorts for the 18 major chronic diseases. }
\begin{tabular}{p{0.1cm}p{2.2cm}p{1cm}p{1.2cm}p{.8cm}p{1cm}}
\toprule
                  \# &  Chronic Disease                                       & Cohort size  & \centering Dominant age group  & \multicolumn{1}{c}{male\%} & white\% \\
                       \midrule
1                      & Hypothyroidism                     & 13,088 & 50-59           & 37.66                     & 72.76  \\
2                      & Stroke                             & 1,698  & 70+             & 50.41                     & 71.97  \\
3                      & Alzheimer's/ Dementia              & 6,138  & 70+             & 42.66                     & 71.01  \\
4                      & Anemia                             & 8,976  & 50-59           & 38.41                     & 64.66  \\
5                      & Asthma                             & 9,180  & 50-59           & 43.15                     & 72.02  \\
6                      & AFib                               & 1,718  & 70+             & 51.98                     & 78.75  \\
7                      & Cardiac                            & 8,156  & 60-69           & 51.80                     & 72.81  \\
8                      & BPH                                & 4,514  & 60-69           & 71.64                     & 72.71  \\
9                      & CKD                                & 15,806 & 50-59           & 47.34                     & 66.94  \\
10                     & COPD                               & 6,088  & 60-69           & 45.63                     & 76.38  \\
11                     & Cancer                             & 3,462  & 60-69           & 46.71                     & 72.93  \\
12                     & Depression                         & 12,458 & 50-59           & 38.91                     & 71.99  \\
13                     & Diabetes                           & 24,324 & 50-59           & 45.63                     & 67.16  \\
14                     & Osteoporosis / Hip fracture        & 3,972  & 60-69           & 26.16                     & 73.51  \\
15                     & Hyperlipidemia                     & 55,620 & 50-59           & 43.88                     & 66.69  \\
16                     & Hypertension                       & 48,276 & 50-59           & 44.36                     & 67.1   \\
17                     & Obesity                            & 32,608 & 50-59           & 43.26                     & 66.53  \\
18                     & Arthritis                          & 8,588  & 60-69           & 41.69                     & 71.29 \\
\bottomrule
\end{tabular}

\label{table:2}
\vspace{1ex}
\tiny
AFib: Atrial Fibrillation, BPH: Benign Prostatic Hyperplasia, CKD: Chronic Kidney Disease, COPD: Chronic Obstructive Pulmonary Disease
\end{table}
\subsection{Study Variables}
Aiming to study the independent role of longitudinal obesity patterns as indicated by BMI trajectories (inputs) on each of the 18 diseases’ incidences (outputs), we have defined a series of newly engineered features to be used as the input to our models. Specifically, we have defined a series of features that capture the trend, seasonality, cyclic nature, and irregularity in the BMI trajectories. The goal of defining these new features is to capture meaningful temporal patterns, without losing the interpretability of the models (a concern with using other methods such as auto-regressive, moving average, or deep learning  to analyze temporal patterns). Additionally, these newly engineered features are directly defined to capture the findings in identifying obesity studies \cite{RN114, RN116, RN113, RN136}. Before defining the new variables, consider a time series $X_i$ that contains $(x_v,t_v )$, for $v=1,2,\dots, V$, where $V$ shows the total number of visits for the $i$-th patient, $x_v$ represents the BMI value at visit $v$, and $t_v$ represents elapsed number of months after the first visit and the visit $v$. We omit index $i$ from the equations since every measurement is done for all patients independently. 

\subsubsection{Weighted Mean}
Simple averaging of the BMI readings cannot fully capture the true average of the irregular time series. There needs to be a mechanism to penalize or rectify large gaps between measurements and therefore the time interval between consecutive measurements can be considered as the weight \cite{RN156}. To capture the overall nature of the BMI trajectories, the weighted mean BMI, shown as $\Bar{X}_{w}$, was calculated using the reciprocals of the time differences between the consecutive visits $v$ and $v-1$ as weight $w_v$.
\begin{equation*}
\small
    \overline{X}_{w}=\sum_{v=1}^{V} {w_v . x_v} / \sum_{v=1}^{V} {w_v},  \; \text{where}, w_v = 1 / (t_v - t_{v-1})
\end{equation*}
\subsubsection{Trend}
The first derivative operation captures the speed of value changes of a certain function. When we have a discrete function instead of a continuous one, we cannot simply use the first derivative, rather an analytical solution can be achieved using the weighted technique. To do this, the Trend $T$ of a BMI trajectory was calculated using the weighted mean of the differences between the BMI values in consecutive visits:
\begin{equation*}
\small    T=\left.\sum_{v=1}^V w_v . (x_v - x_{v-1}) \right/ \sum_{v=1}^V w_v,
\end{equation*}
where, $x_1=x_0$, and $w_v$ is the same as what was defined for the weighted mean. This is the first derivative of the time series scaled using the weight. Techniques similar to this have been used to study other diseases \cite{RN136}. 
\subsubsection{Normalized up-count and down-count}
To elicit the duration of the weight gain (or loss), we calculated the normalized up-rise counts and downfall counts of BMI values in the trajectory. This was done by adding the number of up-rises (or downfalls) of the BMIs in the trajectory and then dividing the result by the number of visits for a patient. This value captures the well-known weight-cycling patterns in individuals' body weights; a phenomenon heavily studied in obesity research \cite{RN174, RN176}. Accordingly, the normalized up-rise count  ($\widetilde{UC}$) and normalized downfall count  ($\widetilde{DC}$) are calculated as follows: 
\begin{equation*}
\small
    \widetilde{UC}={\sum_{v=1}^V U(x_{\Delta v})}/V,
\small
    \text{where } U(x_{\Delta v}) = 
    \begin{cases}
    1,      & \text{if } x_v>x_{v-1}\\
    0,      & \text{otherwise}
    \end{cases}
\end{equation*}
\begin{equation*}
\small
    \widetilde{DC}={\sum_{v=1}^V D(x_{\Delta v})}/V,
\small
    \text{where } D(x_{\Delta v}) = 
    \begin{cases}
    1,      & \text{if } x_v<x_{v-1}\\
    0,      & \text{otherwise}
    \end{cases}
\end{equation*} 
\subsubsection{Maximum BMI}
The maximum BMI value recorded throughout the visit periods was also included as a separate variable. Multiple other studies have shown the importance of including maximum BMI values, for evaluating the association of weight history with mortality \cite{RN114, RN110, RN113}.
\begin{equation*}
\small    BMI_{max}=\max_{v \in V} x_v
\end{equation*} 
\subsubsection{The maximum change in BMI}
The maximum amount of change between two consecutive BMI values over the visits was included to capture the sudden large changes (spikes) in the BMI values that may occur in lieu of steady adjustments of the body weight.
\begin{equation*}
\small    BMI_{max\Delta}=\max_{v \in V} (x_v-x_{v-1})
\end{equation*}
\subsubsection{Start and ending BMI categories}
Standard BMI categories (underweight, normal, overweight, obese) at the beginning and end of the trajectory were included as two separate categorical variables: $BMI_{cs}$ and $BMI_{ce}$. The BMI categories were identified using the US Centre for Disease Control and Prevention (CDC) definitions \cite{RN153}. These two features will also help to capture a simplified picture of the transition of BMI throughout the available timeframe. 
\subsubsection{Median}
The central tendency of the BMI trajectory was recorded using the median of the BMIs throughout the visits.

\subsection{Subtyping (clustering) process}
Formally, in our study, we identify a set of $K$ clusters ($C_1, \dots, C_K$) in each cohort, using the nine variables introduced above. We use the two terms subtyping and clustering interchangeably, as one is more commonly used in biomedical and another in technical settings. The clustering process is repeated for all 18 cohorts, separately. We have applied the $k$-means clustering algorithm on the cohorts. As this was an unsupervised process (meaning the true clusters were not known a priori), we have selected the optimal number of clusters using the elbow method (iteratively minimizing the distance between the cluster centroids and the members) \cite{RN163}. 
To apply $k$-means to a given cohort $X$ containing $n$ multidimensional data points and $K$ clusters to be formed, the Euclidean distance was selected as the similarity index for the clustering aiming to minimize the sum of the squares of the distances between the cluster centers and the data points, as shown below,
\begin{equation*}
\small 
    d=\sum_{k=1}^K \sum_{i=1}^n \|(x_i-u_k)\|^2
\end{equation*}
where $u_k$ represents the $k$-th cluster's center, and $x_i$ represents the $i$-th data point in the corresponding cohort. The quality of the fit of the formed clusters was measured using the Silhouette and Calinski-Harabasz scores \cite{RN107}. Besides the $k$-means method, we have also experimented with a series of traditional (hierarchical: agglomerative with different linkage mechanisms) and modern (deep learning: Deep Temporal Clustering [DTC]\cite{RN134}) clustering methods on the cohorts. Our experiments using various settings showed that $k$-means was able to achieve the best clusters with the least overlapping among the clusters. $K$-means performed best while clustering with Ward linkage and with Complete linkage, and DTC were comparable in cases. We have added the evaluations of the clusters generated by the other methods in the supplementary materials\footnote{\label{note1}https://github.com/healthylaife/athenaHealth} provided in the GitHub repository. Using several t-SNE (t-distributed stochastic neighborhood embedding) plots, we visualized the clusters, and ill-performing clustering methods yielded non-uniformly distributed clusters. In the supplementary materials\textsuperscript{\ref{note1}}, we also provided an example of how clustering was contrasted among different methods.

After generating the clusters for each cohort, we have identified the BMI trends (the overall signal shapes) for each cluster. The BMI trajectories did not have equal lengths, as different individuals had different numbers of visits. We used the $k$-shape algorithm \cite{RN193} to generate a unified shape for the batches of same-length sequences within each cluster. $K$-shape clusters the input same-length trajectories based on their shapes and returns a unified shape for the input trajectories. Then, we combined the unified shapes within the cluster using the time-series $k$-means algorithm \cite{RN194} to generate a final unified sequence representing the whole cluster. The time-series $k$-means uses DTW (Dynamic Time Warping) \cite{RN195} as the distance measure to find the similarities between the sequences. Time-series $k$-means can handle unequal sequences but is computationally very expensive. Therefore, we initially reduced the number of trajectories using the $K$-shape, by combining the equal-length sequences. 

\section{Results}
Before presenting the main results, we first present the results of a series of experiments to verify the relevance of the nine engineered  features for subtyping the cohorts using these features’ values. To this end, we created a series of classifiers: Random Forest, Support Vector Machine (SVM), Decision Tree, Logistic Regression, Gradient Boosting Tree, eXtreme Gradient Boosting Tree (XGBoost) \cite{RN75}, in which the engineered features were used as the input and the disease incidence as the output. This is the same input-output setup as what is used for the clustering. We performed parameter tuning on the classifiers and ran 5-fold cross-validation. Table \ref{table:3} shows the mean accuracy and area under the receiver operating characteristic (AUC)  for each of the individual disease prediction models using these classifiers. Only 8 out of 18 diseases, which are identified as significant in our clustering analysis (discussed later in this section), are shown in Table \ref{table:3}. In this table, the reported results can be compared to a random classifier baseline that should receive 0.5 for both reported measures in all cohorts. 


\begin{table*}[ht]
\tiny
\setlength{\tabcolsep}{2pt}
\caption{Evaluation of the relevance of the engineered features using different classifiers for predicting disease incidence. Mean ($\pm 95$\% CI: Confidence Interval) accuracy and AUC are shown.}
\begin{tabular}{p{1.0cm}|p{1.1cm}|p{1.4cm}|p{1.1cm}|p{1.4cm}|p{1.1cm}|p{1.4cm}|p{1.1cm}|p{1.4cm}|p{1.1cm}|p{1.4cm}|p{1.1cm}|p{1.4cm}} 
\hline
\textbf{Disease}                   & \multicolumn{2}{|c|}{\textbf{RF}}               & \multicolumn{2}{|c|}{\textbf{SVM}}     & \multicolumn{2}{|c|}{\textbf{DT}}     & \multicolumn{2}{|c|}{\textbf{Reg}}   & \multicolumn{2}{|c|}{\textbf{GBT}}        & \multicolumn{2}{|c}{\textbf{XGBT}}              \\ 
\cline{2-13}
                          & ACC  & AUC         & ACC  & AUC          & ACC  & AUC          & ACC  & AUC          & ACC  & AUC          & ACC  & AUC           \\ 
\hline
Diabetes                  & $71.63 \pm 0.5$     & $0.782 \pm 0.004$ & $71.1 \pm 0.3$               & $0.755 \pm 0.004$  & $68.72 \pm 0.2$             & $0.736 \pm 0.003$  & $71.36 \pm 0.3$              & $0.758 \pm 0.003$  & $75.45 \pm 0.2$              & $0.798 \pm 0.002$  & $75.11 \pm 0.2$               & $0.753 \pm 0.006$ \\ \hline
Cancer                    & $71.03 \pm 2.1$            & $0.776 \pm 0.008$ & $68.83 \pm 1.7$              & $0.731 \pm 0.019$ & $66.87 \pm 1.9$            & $0.675 \pm 0.019$ & $68.38 \pm 1.6$            & $0.732 \pm 0.018$ & $73.63 \pm 1.1$             & $0.789 \pm 0.013$ & $76.47 \pm 1.6$              & $0.789 \pm 0.016$  \\ \hline
BPH                       & $71.8 \pm 1.7$            & $0.799 \pm 0.012$ & $73.08 \pm 1.2$              & $0.772 \pm 0.009$ & $69.07 \pm 1.3$            & $0.698 \pm 0.013$ & $73.93 \pm 0.9$             & $0.78 \pm 0.009$ & $75.52 \pm 1.6$             & $0.81 \pm 0.016$ & $79.2 \pm 1.18$              & $0.789 \pm 0.022$  \\ \hline
Osteoporosis/ Hip Fracture & $72.38 \pm 1.6$            & $0.804 \pm 0.013$ & $71.75 \pm 1.8$              & $0.789 \pm 0.019$ & $68.05 \pm 1.6$            & $0.683 \pm 0.014$ & $72.13 \pm 2.1$             & $0.79 \pm 0.02$ & $76.08 \pm 2.2$             & $0.843 \pm 0.017$ & $76.58 \pm 0.6$              & $0.767 \pm 0.057$  \\ \hline
Hypertension              & $71.12 \pm 0.4$            & $0.78 \pm 0.004$ & $70.36 \pm 0.4$              & $0.747 \pm 0.005$ & $68.68 \pm 0.3$            & $0.742 \pm 0.002$ & $70.46 \pm 0.4$             & $0.752 \pm 0.003$ & $74.35 \pm 0.2$             & $0.791 \pm 0.005$ & $74.71 \pm 0.4$              & $0.747 \pm 0.005$  \\ \hline
Alzheimer’s/ Dementia     & $66.55 \pm 1.2$            & $0.725 \pm 0.019$ & $65.57 \pm 1.1$              & $0.712 \pm 0.015$ & $61.4 \pm 1.9$            & $0.609 \pm 0.019$ & $65.7 \pm 1.4$             & $0.713 \pm 0.015$ & $68.68 \pm 1.2$             & $0.754 \pm 0.017$ & $69.39 \pm 0.9$              & $0.725 \pm 0.004$  \\ \hline
Stroke                    & $68.08 \pm 2.2$            & $0.751 \pm 0.022$ & $70.49 \pm 2.4$              & $0.748 \pm 0.017$ & $65.9 \pm 1.2$            & $0.663 \pm 0.009$ & $70.38 \pm 2.7$             & $0.753 \pm 0.019$ & $72.44 \pm 1.2$             & $0.769 \pm 0.02$ & $71.81 \pm 0.8$              & $0.725 \pm 0.004$  \\ \hline
Obesity                   & $88.25 \pm 0.3$            & $0.919 \pm 0.002$ & $86.59 \pm 0.3$             & $0.907 \pm 0.005$ & $85.34 \pm 0.6$            & $0.905 \pm 0.005$    & $87.25 \pm 0.3$             & $0.909 \pm 0.004$ & $90.34 \pm 0.2$             & $0.928 \pm 0.003$ & $90.48 \pm 0.2$             & $0.905 \pm 0.008$  \\ 
\hline
\end{tabular}
\label{table:3}
\vspace{1ex}
\tiny
RF: Random Forest, SVM: Support Vector Machine, DT: Decision Tree, Reg: Logistic Regression, GBT: Gradient Boosting Tree, XGBT: Extreme Gradient Boosting Tree. \\
ACC: Accuracy. AUC: area under the receiver operating characteristic curve.
\end{table*}

\begin{centering}
\begin{figure*}[t!]
\includegraphics[width=\linewidth,height=12cm]{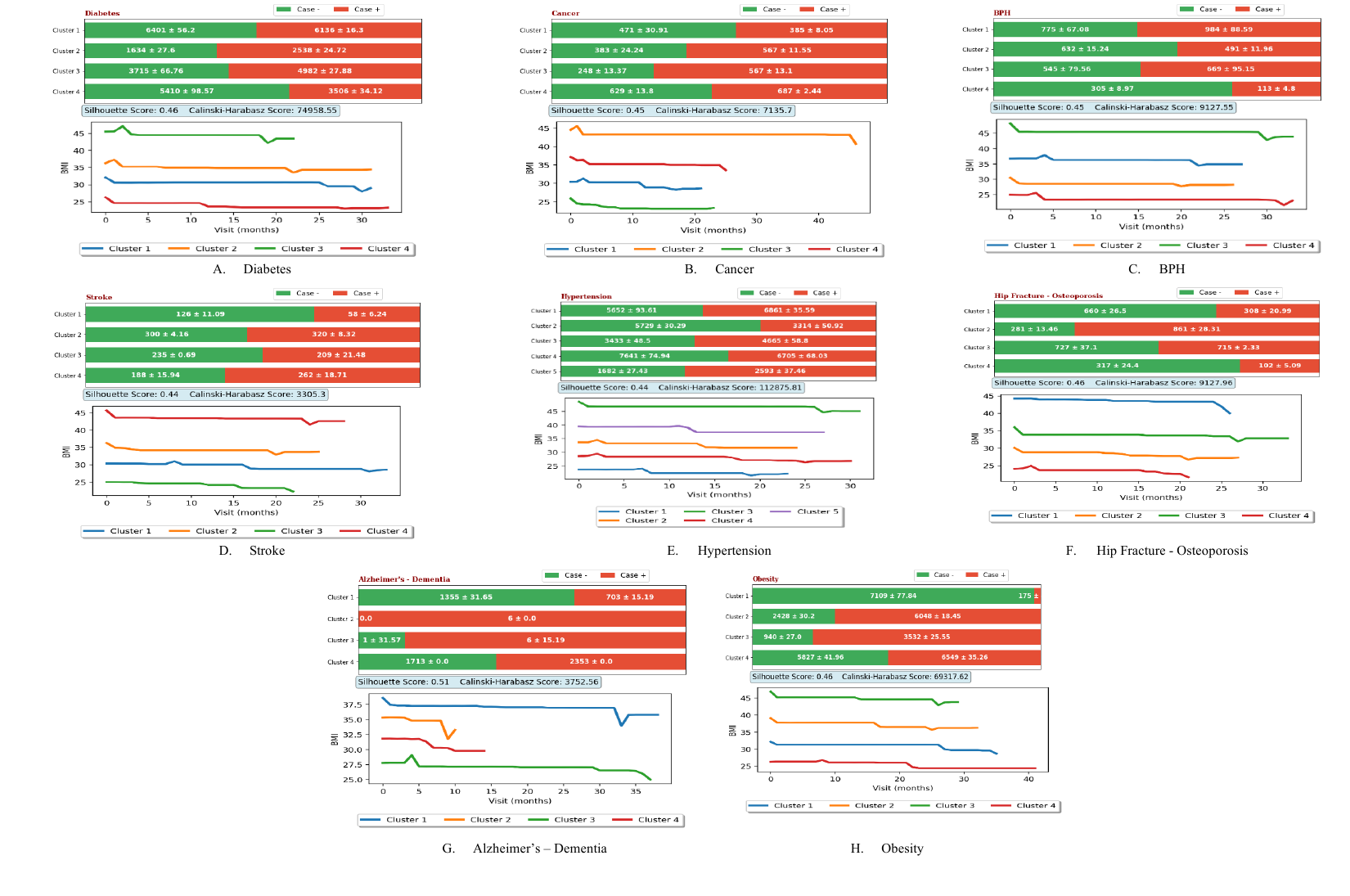} 
\caption{Distribution of the clusters for the 8 chronic diseases and the overall BMI trends (the shapes of signals) for each cluster. The top chart (horizontal bars) in each sub-figure shows how the positive (those with the disease incidence) cases and negative cases (those without the disease incidence) are distributed within the identified subgroups (clusters). The bottom chart (vertical bars) shows the BMI trends corresponding to each cluster. The mean number of individuals ± (std) is shown for each cluster. }
\label{fig:image1}
\end{figure*}
\end{centering}

For the main experiments of this study, we then applied the presented clustering method on the 18 cohorts. The cluster analyses revealed the existence of noticeable clustered patterns in 8 of the 18 cohorts, as determined by the cluster evaluation metrics (Silhouette and Calinski-Harabasz scores \cite{RN164}) and visual confirmation (t-SNE plots \cite{RN173}). We provide several t-SNE visualizations of multiple clustering techniques in our GitHub repo. A higher similarity between the members of each cluster or a higher difference between the members of separate clusters leads to higher Silhouette and Calinski-Harabasz scores. Fig. \ref{fig:image1} shows the clusters identified in the 8 noticeable cases. The absence of distinct patterns among the remaining 10 chronic diseases may indicate the absence of BMI-related subtypes for each chronic disease or may be due to the insufficient number of positive cases for the diseases in our data (such as AFib). We calculated the relative risk \cite{RN137} of the clusters showing either positive-case domination or negative-case domination to compare the population in those clusters with the remaining ones.

Having the identified subgroups for the 8 diseases, we then study the disparities in each subgroup (of each of the 8 cohorts) with respect to the demographic and socioeconomic factors, as well as the differences in physiological measurements presented. We validated the differences in the variables using statistical tests and report the results in Table \ref{table:4}. For the significance analysis, we used Chi-squared tests \cite{RN165} for categorical variables and one-way ANOVA \cite{RN166} for continuous variables. Variables with $p$-value less than 0.5 are marked with ** and those with $p$-value less than 0.1 with *. In Table \ref{table:5}, we present a cluster distribution compared to the asymptotic (devoid of any disease diagnosis) population. In this table, for the categorical variables, the number of patients within that category is shown along with the continuous variables, presented with the mean value within the cluster and the standard deviation. We refer readers to our online GitHub repo for similar cluster distributions among each of the cohorts, as well as our code. 

Besides the 18 separate cohorts, we also apply the $k$-means clustering method on a combined cohort ($N$ = 146,857) of patients with at least one of the 18 diseases diagnosed (the cohort containing the patients in any of the 18 cohorts). The goal of this part of our experiments was to compare the patterns in the subgroups identified in patients with any chronic disease incidence to the asymptotic cohort (those without a diagnosis of the 18 major chronic diseases). Table \ref{table:5} shows the differences among the four clusters formed on the combined cohort, contrasted against the asymptotic cohort, as determined by the cluster evaluation metrics. In this table, each continuous variable is presented with mean and standard deviation while categorical variables are presented, with the total number of patients in those categories.

\begin{table}
\scriptsize
\setlength{\tabcolsep}{1pt}
\centering
\caption{Variable difference  among  the  subgroups of  8 chronic disease cohorts (* and ** represent P$<$0.05 and P$<$0.01 respectively). 1) diabetes, 2)cancer, 3)BPH, 4)Hip fracture/Osteoporosis, 5)Hypertension, 6) Alzheimer’s/ Dementia, 7)Stroke, 8) Obesity. }
\begin{tabular}{p{2.4cm}p{0.7cm}p{0.7cm}p{0.7cm}p{0.7cm}p{0.7cm}p{0.7cm}p{0.7cm}p{0.7cm}}
\hline
~&\textbf{1}&\textbf{2}&\textbf{3}&\textbf{4}&\textbf{5}&\textbf{6}&\textbf{7}&\textbf{8}
\\ \hline
\textbf{Age}            & **       & *      & ~   & *                          & **           & ~                     & ~      & **       \\ 
\textbf{Income}         & ~        & ~      & ~   & ~                          & **           & ~                     & ~      & ~        \\ 
\textbf{Insurance}      & *        & ~      & ~   & ~                          & *            & ~                     & ~      & *        \\ 
\textbf{Race/ethnicity} & *        & ~      & *   & ~                          & *            & ~                     & ~      & *        \\ 
\textbf{Residence}      & ~        & ~      & ~   & ~                          & ~            & ~                     & ~      & ~        \\ 
\textbf{Gender}         & ~        & ~      & ~   & ~                          & **           & ~                     & ~      & *        \\ 
\textbf{HbA1c}          & **       & **     & **  & **                         & **           & **                    & **     & **       \\ 
\textbf{Systolic BP}    & **       & **     & **  & **                         & **           & **                    & **     & **       \\ 
\textbf{Diastolic BP}   & **       & **     & **  & **                         & **           & **                    & **     & **       \\ 
\textbf{LDL}            & **       & ~      & ~   & ~                          & **           & **                    & ~      & **      \\ \hline
\end{tabular}
\label{table:4}
\end{table}

\section{Discussion}
We separately discuss the results of our clustering experiments for each of the 8 diseases, which were identified as the diseases with recognizable patterns in the identified clusters. Specifically, we report the unique characteristics of each of the clusters with distinct positive or negative cases (not all clusters), in terms of demographic, socioeconomic, and physiological measurements, as well as the relative risks of disease incidence. For the comparison of the physiological measurements, we consider normal HbA1c levels for (type 2) diabetes as normal, when below 5.7\%, pre-diabetes 5.7\%-6.4\%, and diabetes above 6.4\% (therefore, any measurement above 6.4\% is considered dangerous) as directed by WHO (World Health Organization) \cite{RN307}. For the LDL cholesterol, we follow CDC’s definition, considering LDL levels of above 100 mg/dL as bad for health, which needs to be consulted with a health provider \cite{RN149}. For the blood pressure, we follow CDC guidelines outlining normal blood pressure levels being less than 120 mmHg systolic and 80 mmHg diastolic \cite{RN150}.

\begin{table}[ht]
\setlength{\tabcolsep}{1pt}
\tiny
\centering
\caption{Cluster analysis of the four clusters from the combined cohort (containing the combination of the individuals in any of the 18 cohorts).  The asymptotic cohort includes individuals without a diagnosis of the 18 major chronic diseases. Each cell contains the number of individuals with the characteristics (described at row level) in that cluster (defined by the column).}
\begin{tabular}{p{1.1cm}p{1.6cm}p{1.1cm}p{1.1cm}p{1.1cm}p{1.1cm}p{1.1cm}} 
\hline
~              & \textbf{Description}            & \textbf{Asymptotic}     & \textbf{Cluster 1}      & \textbf{Cluster 2}      & \textbf{Cluster 3}     & \textbf{Cluster 4}       \\ 
\hline
\textbf{Age}            & $<$30                     & 3,831           & 948            & 1,197           & 735           & 1,183            \\ 
~              & 30-39                  & 4,358           & 2,795           & 2,466           & 1,537          & 3,258            \\ 
~              & 40-49                  & 4,329           & 7,848           & 4,279           & 3,866          & 7,118            \\ 
~              & 50-59                  & 3,530           & 14,111          & 5,144           & 8,218          & 10,493           \\ 
~              & 60-69                  & 1,771           & 16,443          & 3,839           & 11,123         & 10,004           \\ 
~              & 70+                    & 782            & 11,406          & 1,068           & 12,899         & 4,879            \\ 
\hline
\textbf{Income}         & High                   & 5,777           & 9,718           & 2,505           & 8,305          & 6,090            \\ 
~              & Low                    & 7,685           & 28,976          & 10,598          & 19,927         & 20,617           \\ 
~              & Medium                 & 5,139           & 14,857          & 4,890           & 10,146         & 10,228           \\ 
\hline
\textbf{Insurance}      & Commercial             & 13,003          & 22,415          & 9,269           & 12,615         & 17,727           \\ 
~              & Medicaid               & 2,137           & 2,820           & 2,115           & 2,057          & 2,694            \\ 
~              & Medicare               & 2,353           & 25,875          & 5,869           & 22,266         & 14,844           \\ 
~              & Other                  & 60             & 58             & 15             & 32            & 44              \\ 
~              & Self-pay               & 1,048           & 2,383           & 725            & 1,408          & 1,626            \\ 
\hline
\textbf{Race/ethnicity} & Black / African Amer. & 1,560           & 6,947           & 3,618           & 4,210          & 5,782            \\ 
~              & Hispanic / Latino      & 3,279           & 7,082           & 1,777           & 4,567          & 4,395            \\ 
~              & Other                  & 1,799           & 2,166           & 290            & 3,232          & 1,035            \\ 
~              & White                  & 11,963          & 37,356          & 12,308          & 26,369         & 25,723           \\ 
\hline
\textbf{Residence}      & Metro                  & 16,047          & 42,561          & 14,253          & 31,110         & 29,061           \\ 
~              & Metro-adjacent         & 1,887           & 7,531           & 2,701           & 4,825          & 5,520            \\ 
~              & Rural                  & 667            & 3,459           & 1,039           & 2,443          & 2,354            \\ 
\hline
\textbf{Gender}         & Female                 & 11,477          & 24,630          & 11,638          & 19,740         & 19,783           \\ 
~              & Male                   & 7,124           & 28,921          & 6,355           & 18,638         & 17,152           \\ 
\hline
\textbf{HbA1C}          & Mean (std)             & 5.63 (0.68)    & 6.8 (1.41)     & 7.01 (1.49)    & 6.62 (1.42)   & 6.96 (1.45)     \\ 
\hline
\textbf{LDL}            & Mean (std)             & 104.22 (27.92) & 98.78 (29.46)  & 99.68 (28.22)  & 98.19 (29.72) & 99.12 (28.93)   \\ 
\hline
\textbf{Systolic BP}    & Mean (std)             & 121.26 (11.76) & 130.52 (11.66) & 132.98 (11.42) & 128.78 (12.7) & 131.49 (11.38)  \\ 
\hline
\textbf{Diastolic BP}   & Mean (std)             & 75.52 (7.64)   & 77.37 (7.49)   & 80.16 (7.52)   & 75.04 (7.74)  & 78.78 (7.46)    \\ 
\hline
\textbf{Total}          & ~                      & 18,601          & 53,551          & 17,993          & 38,378         & 36,935           \\
\hline
\end{tabular}
\label{table:5}
\end{table}

\textbf{Diabetes} - As obesity (and therefore the BMI pattern) is a primary risk factor for diabetes \cite{RN117, RN139}, one would expect to also see stronger patterns identified in our cluster analysis of the diabetes cohort. As shown in Fig. \ref{fig:image1}A, the four clusters identified in this cohort show stronger patterns compared to the other clusters. Cluster 2 has 60.83\% diabetic patients and cluster 3 contains 57.28\% of them. Looking further into the clusters, it is found that the average health conditions (blood pressure, HbA1c, LDL) of these individuals are considerably higher, in fact, highest for cluster 2 (mean HbA1c 6.9\%, systolic BP 132.78 mmHg, diastolic BP 81.24 mmHg, and LDL 104.53 mg/dL). Observing the trend of BMI in these two clusters, we find the trend shows less decline and stays above 30. The remaining two clusters show a decline in BMI towards the end. The dominant age range in clusters 2 and 3 is 40-59 (52.3\% of the total cluster population) and a significantly higher percentage (18.96\%) are Black or African American. Cluster 2 also has a significant number (19.42\%) of individuals from the age group 30-39, whereas cluster 3 has a relatively higher number of older adults (28.13\%) from the age group 50-59. Interestingly, cluster 4, which contains around 60\% of healthy individuals with relatively healthy physiological measurements, consists of individuals from age 50 and above. Individuals in cluster 4 have a 35\% less risk of diabetes than the individuals of cluster 2 and 31\% less risk of diabetes than the individuals of cluster 3. One of the results of obesity is building up insulin resistance \cite{RN117}. As a result, more than 80\% of cases of diabetes can be attached to obesity, obese men were found to have 7-fold and obese women a 12-fold higher chance to develop diabetes \cite{RN117} respectively. In our study, we found that the two clusters with mostly positive cases consisted of a higher percentage (65.99\% and 55.68\% respectively) of women than men. Current guidelines suggest that screening for diabetes should begin at the age of 45 years, and individuals who are overweight or obese are recommended for diabetes screening without any specific age restrictions \cite{RN118}. Our findings also support this guideline by finding most of the positive cases in the diabetes cohort to be in clusters with age groups 40-59. The mean age of the individuals diagnosed with diabetes was found to be 46.3 years in the study by Eberhardt et al. \cite{RN120}, which aligns with our study. In the same study, adults with a diagnosed diabetes had a lower family income, which aligns with the two distinct clusters (2 and 3) in the diabetes cohort that had lower income categories (57.77\% and 53.43\% respectively) compared to other clusters \cite{RN120}.

\textbf{Cancer} – Clusters 1 and 4 in Fig. \ref{fig:image1}B have almost equal distributions of positive and negative cases, but clusters 2 and 3 contain some partiality of the cases.  59.68\% of the individuals are positive cases, in cluster 2. Also, around 30\% of them are from the high-income category, and more than 80\% are in the age groups 50-59, 60-69, and above 70, while having relatively healthy conditions with the lowest measurements in all clusters (mean HbA1c 6.06\%, systolic BP 125.88 mmHg, diastolic BP 74.06 mmHg, and LDL 100.74 mg/dL). On the contrary, cluster 3, with 72.94\% of healthy individuals, does not have healthy measurements. The mean systolic blood pressure is 132.29 mmHg, the mean HbA1c measure is 6.84 and the mean diastolic blood pressure is close to 80 mmHg. A significant amount of these individuals are Black or African American  (17.35\%) with the majority in the low-income category (60.59\%). DeSantis et al. have also found a significant decline in cancer mortality among Black or African American individuals \cite{RN121}. This points to the fact that in our study, the cluster with mostly healthy individuals comprised a good number of Black or African American individuals. Cluster 3 also has fewer older adults compared to the other clusters and comprises mostly individuals from 30 to 69 years of age. There is a 2.21 times higher risk of cancer in cluster 2 compared to cluster 3. The BMI trendlines of cluster 2 show a steady BMI above 40, whereas the trendline of cluster 3 has a gradual decline, staying close to 20.

\textbf{BPH} - Cluster 4 in Fig. \ref{fig:image1}C has 72.97\% negative cases with individuals having overall the worst health conditions (mean HbA1c 6.68\%, systolic BP 131.92 mmHg, diastolic BP 79.24 mmHg, LDL 100.62 mg/dL). Those in this cluster have almost two times higher risk of diagnosing with BPH than the individuals in clusters 1 or in 3. With 56.7\% comprising low-income individuals, this cluster (4) is covered by individuals with low- and mid-income categories (86.13\% jointly) and individuals with Black or African American ethnicity (17.22\%). While obesity is reported to have a strong (positive) association with BPH and lower urinary tract symptoms \cite{RN122}, finding only one cluster with mostly healthy individuals in our study may be due to the role of other covariates also playing a role in the development of BPH. The trendline of cluster 4 shows a steady BMI in the normal range.

\textbf{Stroke} - For the stroke cohort, there are 4 clusters with 2 of them showing a majority in negative or positive cases. In Fig. \ref{fig:image1}D, cluster 1 contains almost 70\% of the negative cases. However, their physiological measurements have the worst health conditions among the individuals from all 4 clusters (mean HbA1c 6.9\%, systolic BP 134.52 mmHg, diastolic BP 80.31 mmHg, LDL 96.56 mg/dL). On the contrary, cluster 4, with almost 60\% positive cases, has better health conditions (mean HbA1c 6.24\%, systolic BP 127.86 mmHg, diastolic BP 74.48 mmHg, and LDL 95.37 mg/dL). These individuals from cluster 4 are mostly aged above 70 and are low-income. The individuals in cluster 1 are 46\% less likely to have a stroke compared to the individuals in cluster 4. As found by Sommer et al. \cite{sommer2018impact}, female patients with less healthy conditions develop less risk of cardiovascular diseases than overweight or obese male patients. Cluster 1 consists of about 60\% female individuals which explains the less likelihood of stroke in the cluster. Cluster 1 shows a steady BMI trend with a gradual decline and cluster 4 shows a very high BMI trend with sharp drops at the beginning and end. In the work of Polivka et al. \cite{polivka2019risks}, we find that the risk of cardiovascular events is associated with abnormal BMI. The U-shaped risk association affects both overweight and underweight. The individuals in cluster 4 with almost 60\% positive cases show the average BMI value is above 42 $\mathrm{kg/{m^2}}$. 

\textbf{Hypertension} - Among the  five clusters in Fig. \ref{fig:image1}E, clusters 2 and 5 show partial saturation of negative and positive patients, respectively. Cluster 2 contains more than 60\% negative cases with the best of the flock health measurements (mean HbA1c 6.02\%, systolic BP 124.22 mmHg, diastolic BP 74.71 mmHg, LDL 103.41 mg/dL). This cluster is composed of individuals from varied age ranges, but the percentage of samples from the age group 50-69 is significantly higher (44.08\%). On the other hand, cluster 5 includes samples from ages 30-59 in higher quantity. 60.65\% of the samples in this cluster  have hypertension as a condition with the worst physiological measurements (mean HbA1c 6.6\%, systolic BP 132.96 mmHg, and LDL 105.53 mg/dL). The composition of these cluster shows distinct variation for the race of Black or African American and the Other category (races other than White or Hispanic/Latino). 8.66\% of the population is Black or African American in cluster 2 against 18.62\% of cluster 5. Also, 12.46\% individuals of cluster 2 are from the “Other” race category against 2.11\% of cluster 5. Cluster 2 also has a higher percentage (30.7\%) of individuals from the high-income category and a lesser percentage (42.01\%) from the low-income category than those of cluster 5 (high-income 14.01\%, low-income 57.61\%). Compared to the individuals in cluster 2, the individuals in cluster 5 have a 1.66 times risk of developing hypertension. Obesity has been found to be one of the important risk factors for hypertension by numerous studies \cite{RN129, RN128}. Our study finds that the prevalence of hypertension in African Americans is greater than the rate found elsewhere \cite{RN130}, while this reference study also found a significant number of positive individuals are of Hispanic origin. The trendlines of clusters 2 and 5 show similarity, however, cluster 2 has a lower BMI trend having more healthy populations and cluster 5 has a higher BMI trend.

\textbf{Hip fracture and Osteoporosis} - Cluster 1 and 4 in Fig. \ref{fig:image1}F contain 68.18\% and 75.66\% negative cases, respectively. The ratio of individuals with Black or African American ethnicity to all ethnicity in these clusters is comparatively higher than the other clusters, 12.09\% and 16.71\% respectively compared to the other clusters’ 4.82\% and 9.22\%. These clusters also contain a significant percentage (53.51\% and 58.71\% respectively) of individuals with low income. However, these individuals have critical health conditions, with cluster 4 having the most critical conditions with higher mean HbA1c (6.7\%) and mean blood pressure measurements (systolic BP 132 mmHg, diastolic BP 79.77 mmHg). There is 58\% and 68\% less risk of hip fracture or osteoporosis in clusters 1 and 4 compared to cluster 2, respectively. Cluster 2 contains around 75\% positive cases, but has comparatively healthy individuals evaluated through physiological measurements (mean HbA1c 5.93\%, systolic BP 125.6 mmHg, diastolic BP 73.65 mmHg, LDL 103.84 mg/dL). This cluster has a significant portion (30.74\%) of its members from the high-income category. The cohort with a hip fracture or osteoporosis shows clustering such that clusters with higher positive cases are comparatively healthy in terms of physiological measurements, and the individuals with worse health conditions have lesser positive cases of hip fracture or osteoporosis. Similar findings have been reported in  previous studies \cite{RN124, RN125}, while there are few studies that show a correlation between obesity and fractures \cite{RN126}. The BMI trendlines of clusters 1 and 4 show extremities on the chart. Cluster 1 has a high BMI trend with a gradual decrease in value whereas cluster 4 has a low BMI trend with a steep decline.

\textbf{Alzheimer’s and Dementia} - From the Alzheimer’s and dementia cohort, we found 4 clusters (Fig. \ref{fig:image1}G); two clusters with a very small number of members (6 and 7 members), potentially referring to some rare cases. Cluster 2, with all of its members being positive, contains older adults in the age range of 60 and above, and 4 of them are from the low-income category. Cluster 3 has 85.71\% positive cases from a similar age group to cluster 2. Both these clusters contain white, black, or African American individuals. The individuals in clusters 2 and 3 are 2.93 and 2.51 times more likely to develop Alzheimer’s or dementia than the individuals in cluster 1. They have the worst physiological measurements, where cluster 2 has the highest HbA1c of 6.9\% and diastolic BP of 77.01 mmHg, whereas cluster 3 has the highest LDL of 99.27 mg/dL and systolic BP of 130.98 mmHg from the individuals in those clusters. Cluster 1 shows a peculiar characteristic among its members where individuals in this cluster also have close to the highest physiological measurements (mean HbA1c 6.78\%, systolic BP 130.56 mmHg, diastolic BP 77.08 mmHg, LDL 97.32 mg/dL), while more than 65\% of them are negative cases. As discussed by Qu et al. \cite{qu2020association}, Pegueroles et al. \cite{pegueroles2018obesity} late-life underweight acts as a positive risk associated with incident dementia, the correlation of which is unlikely to be causal, as weight loss is well-defined to be presented with comorbidities in late-life. Since the individuals in this cluster have a relatively high BMI trend also with most of them being aged 60 and above, they comprise the cluster with more than 65\% negative cases. The other cluster (cluster 4) contains 57.87\% positive cases with relatively better health conditions (mean HbA1c 6.45\%, systolic BP 128.25 mmHg, diastolic BP 73.76 mmHg, LDL 95.3 mg/dL). Cluster 2 shows high BMI trends and cluster 3 shows a lower and steady BMI trend.

\textbf{Obesity} - Not surprisingly, subtyping in the obesity cohort (where  a BMI above 30 is considered a diagnosed disease) led to the most distinct patterns (Fig. \ref{fig:image1}H). Cluster 1 has more than 95\% of the healthy patients, whereas clusters 2 and 3 have 71.35\% and 78.98\%  patients with obesity, respectively. Cluster 4 has almost an equal share of obesity and healthy individuals. Individuals in clusters 2 and 3 have 29.7 and 32.87 times higher risk of being diagnosed with obesity, compared to the individuals in cluster 1, respectively. Moreover, the risk of obesity in cluster 3 is 1.11 times as high as the risk in cluster 2. Cluster 1 with a higher number of healthy individuals comprises individuals of age more than 70 years and 25.41\% of the total population of the cluster. However, this cluster has less proportion of black, African American, or Hispanic/Latino individuals (10.23\% of the total population of the cluster). Also, the mean HbA1c, blood pressure, and LDL are the lowest (6.272\%) in this cluster compared to others. If we characterize cluster 1 by individuals in the age range of 50 and above with the best health conditions, then cluster 2 can be characterized by young and middle-aged individuals with high LDL values (mean 105.85 mg/dL) and moderately worse health conditions. Cluster 3 can be characterized by young-adult individuals of the age range of less than 30 to 59 with high blood pressure and HbA1c levels (mean 6.49). The mean systolic and diastolic pressure of the individuals in cluster 3 are 132.27 mmHg and 81.47 mmHg, respectively. Cluster 1 has most of the healthy individuals show a steady declining BMI trend around the mark of 30. On the other hand, clusters 2 and 3 have high BMI trends with a sharp decline in the beginning and end.

\textbf{Combined cohort} - Besides the clusters identified from the 18 cohorts, we have also studied the patterns discovered from clustering the combined cohort (patients with any of the 18 diseases) compared to the asymptotic individuals (those without any of the 18 diagnoses), as shown in Table \ref{table:4}. As anticipated, the physiological measurement values are better in the asymptotic cluster, where the overall mean of Hemoglobin A1c, systolic and diastolic blood pressures are lower than the diseased clusters. The LDL measurements are similar (98.78, 99.68, 98.19, and 99.12 mg/dL respectively) in the diseased clusters; however, it is higher in the asymptotic one (104.22 mg/dL). The asymptotic cluster contains mostly younger individuals, dominantly the age groups 30-39 (23.43\%) and 40-49 (23.27\%). This cluster has a higher composition (31.06\%) of patients with high income, whereas the diseased clusters have many patients with low incomes (54.11\%, 58.9\%, 51.92\%, and 55.82\% respectively). There is also significant disparity among patients of different races or ethnicity. Though all the clusters are predominantly white, the second largest population (cluster 3) shows a higher composition of patients from black or African American ethnicity. Almost 18\% of the asymptotic cluster are from the Hispanic/Latino race with fewer black or African American patients (8.39\%). All diseased clusters contain more than 10\% of black or African Americans (12.97\%, 20.11\%, 10.97\%, and 15.65\%) respectively in clusters 1-4. Finally, those from the rural areas comprise only 3.59\% of the asymptotic cluster, but around 6\% in every other cluster, showing the considerable differences between the urban and rural areas. 

Among the clusters with mostly diseased individuals, clusters 2 and 4 show worse physiological measurements. Cluster 2 has a mean HbA1c of 7.01\%, systolic BP of 132.98 mmHg, diastolic BP of 80.16 mmHg, and LDL of 99.68 mg/dL. In these clusters the predominant diagnosed disease is obesity. Of the  patients with obesity, 30\% of them are in cluster 2 and around 40\% are in cluster 4. Cluster 1 has  a mixture of diseases among the members, however, the number of patients with hip fractures or osteoporosis is considerably less in it. Only 21.59\% of the patients with this condition are in this cluster. Cluster 3 is predominantly the patients with hip fracture or osteoporosis (70.6\% of the total population with this condition) and with Alzheimer’s or dementia (62.61\% of the total population with this condition). 

\textbf{Limitations} - Several limitations of our study should be noted. In our study, we did not create separate cohorts for males and females to avoid reducing the size of the cohorts. Still, we study the role of gender in our significance analysis (Table \ref{table:4}). Also, we used the $k$-means clustering technique for generating the cluster and tested some other clustering techniques (agglomerative and deep learning). With more data being available and accessible in the future, techniques that automatically identify the temporal features (such as recurrent neural networks \cite{RN184} and transformers \cite{RN186}) may reveal more hidden patterns. These advanced machine learning, however, often suffer from interpretability concerns \cite{RN190, RN187, RN188, RN189}. Our method relies on a series of interpretable and clinically informed features. Our study was centered around the hypothesis that BMI trends alone can estimate the incidence of some of the major chronic diseases. Each of the chronic diseases studied here results from a complex interaction between many covariates, where BMI patterns (or obesity status) are one of those. While our study targets the association between temporal BMI patterns and major chronic diseases, it does not explicitly target the role of these BMI patterns in the incidence of those diseases. An even larger dataset (than our very large dataset) would allow the creation of subgroups  that are large enough for this type of analysis (i.e., subgroups with distinct characteristics of incidence of a chronic disease following a sufficient number of BMI recordings). While our study is unique in its approach of considering only BMI trends, the role of such trends can be also studied in conjunction with other covariates. 
\section{Conclusion}
In this work, we have studied the independent role of BMI trajectories (and obesity status) in the onset of 18 major chronic diseases using nine engineered features that we proposed. We found that BMI patterns can be effectively utilized to find the characteristics of the individuals at risk of developing eight of those chronic diseases. We have identified specific subgroups (clusters) of individuals at higher risk of developing those eight diseases, along with the characteristics that stand out. Among the strongest patterns we found, the direct relationship of BMI trajectories with diabetes, hypertension, and dementia was re-established by the findings of our study. We showed a subgroup in the cancer cohort with mostly healthy individuals comprised of a good number of black or African American individuals. Additionally, two distinct clusters with either positive or negative cases dominating the cluster have been identified in the hypertension cohort. Studying the BPH cohort suggests there is scope for further investigation to identify the key role players in the development of BPH other than BMI patterns alone. An inverse relation with BMI trajectories in hip fracture and osteoporosis cohort showed that negative case patients turned out to have healthier physiological measurements. All these segregations build up the fact that patient subtyping  based on the BMI trajectory patterns propounds the role of obesity as a factor of the major chronic diseases.

\bibliographystyle{unsrt}
\bibliography{IEEE_JBHI_Dip_BMI.bib}

\end{document}